\documentclass[nohyperref]{article}

\usepackage{amsmath,amsfonts,bm}

\newcommand{\norm}[1]{\left\lVert\,#1\,\right\rVert}

\def\eqref#1{equation~\ref{#1}}

\def\1{\bm{1}}

\DeclareMathAlphabet{\mathsfit}{\encodingdefault}{\sfdefault}{m}{sl}
\SetMathAlphabet{\mathsfit}{bold}{\encodingdefault}{\sfdefault}{bx}{n}

\usepackage{microtype}
\usepackage{graphicx}
\usepackage{subfigure}
\usepackage{booktabs} %

\usepackage{hyperref}

\usepackage[accepted]{icml2023}

\usepackage{amsmath}
\usepackage{amssymb}
\usepackage{mathtools}
\usepackage{amsthm}

\usepackage[capitalize,noabbrev]{cleveref}

\theoremstyle{plain}

\theoremstyle{definition}

\theoremstyle{remark}

\usepackage{xspace}
\usepackage{lipsum}
\newcommand{\p}[1]{\vspace{1mm}\noindent\textbf{#1}}

\newcommand{\ie}{\textit{i}.\textit{e}.}
\newcommand{\eg}{\textit{e}.\textit{g}.}

\usepackage{pifont}
\newcommand{\cmark}{\ding{51}}%
\newcommand{\xmark}{\ding{55}}%
\usepackage{tabularx}
\usepackage{multirow}

\icmltitlerunning{Language Quantized AutoEncoders: Towards Unsupervised Text-Image Alignment}
\newcommand{\ours}{{LQAE}\xspace}

\begin{document}

\twocolumn[
\icmltitle{
Language Quantized AutoEncoders: \\ Towards Unsupervised Text-Image Alignment}

\begin{icmlauthorlist}
\icmlauthor{Hao Liu}{}
\icmlauthor{Wilson Yan}{}
\icmlauthor{Pieter Abbeel}{} \\
University of California, Berkeley
\end{icmlauthorlist}

\icmlcorrespondingauthor{Hao Liu }{hao.liu@cs.berkeley.edu}

\icmlkeywords{Machine Learning, ICML}

\vskip 0.3in
]

\printAffiliationsAndNotice{University of California, Berkeley}  %

\begin{abstract}

Recent progress in scaling up large language models has shown impressive capabilities in performing few-shot learning across a wide range of text-based tasks. However, a key limitation is that these language models fundamentally lack visual perception - a crucial attribute needed to extend these models to be able to interact with the real world and solve vision tasks, such as in visual-question answering and robotics. Prior works have largely connected image to text through pretraining and/or fine-tuning on curated image-text datasets, which can be a costly and expensive process. 
In order to resolve this limitation, we propose a simple yet effective approach called \textbf{L}anguage-\textbf{Q}uantized \textbf{A}uto\textbf{E}ncoder (LQAE), a modification of VQ-VAE that learns to align text-image data in an \emph{unsupervised} manner by leveraging pretrained language models (\eg BERT, RoBERTa). Our main idea is to encode image as sequences of text tokens by directly quantizing image embeddings using a pretrained language codebook. We then apply random masking followed by a BERT model, and have the decoder reconstruct the original image from BERT predicted text token embeddings. 
By doing so, LQAE learns to represent similar images with similar clusters of text tokens, thereby aligning these two modalities without the use of aligned text-image pairs. This enables few-shot image classification with large language models (\eg, GPT-3) as well as linear classification of images based on BERT text features. To the best of our knowledge, our work is the first work that uses unaligned images for multimodal tasks by leveraging the power of pretrained language models.
\vspace{2.0em}
\end{abstract}

\section{Introduction}

Large language models powered by transformers~\citep{ashish2017attention} have achieved impressive results on modeling natural language~\citep[see e.g.][]{brown2020language, ouyang2022training, chowdhery2022palm, zhang2022opt}. 
Notably, they can learn to perform new tasks, such as question answering, chatbots, and machine translation from just a few examples without finetuning. This so called few-shot learning turns out to be competitive with conventional task specific methods in various NLP tasks and is being rapidly adapted to more new tasks and styles of generations.

Despite these impressive capabilities, a key limitation of such large language models is that they cannot `see' the visual world. 
Being able to `see' the world is crucial for many real world applications where processing abundant complex sensory data is a must, such as robotics, visual question answering, and grounding. Such a limitation severely hinders further applicability of large transformer models to a variety of downstream tasks.

Driven by these impressive results for large language models, much of recent work has started to bring text and image together, and leverage these aligned modalities to perform a variety of applications, such as text to image generation~\citep{yu2022scaling, chang2023muse}, open ended classification~\citep{tsimpoukelli2021multimodal, radford2021learning, alayrac2022flamingo}, and image editing~\citep{gal2022stylegan,meng2021sdedit,ramesh2022hierarchical}.
Frozen~\citep{tsimpoukelli2021multimodal} trains a visual encoder to adapt images to pretrained language model for predicting aligned text, and demonstrates few-shot learning in classification and VQA. 

\begin{figure*}[t]
    \centering
    \includegraphics[width=0.95\textwidth,keepaspectratio]{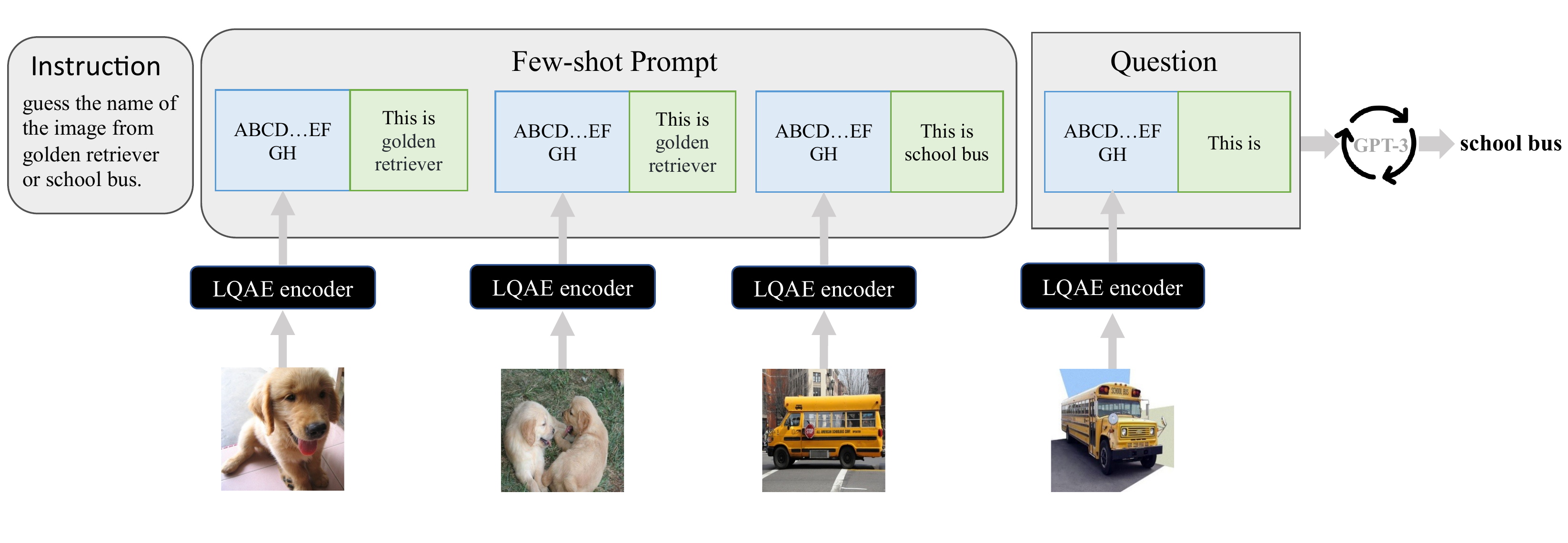}
    \caption{Language Quantization AutoEncoder (\ours) can be used for few-shot image classification by leveraging the in-context learning ability of large language models \eg, GPT-3}
    \label{fig:fewshot_cls_model}
\end{figure*}

However, these works generally require large amounts of aligned data - Frozen is pretrained on Conceptual Captions~\citep{sharma2018conceptual} (3M text-image pairs), and many prior methods use CLIP~\citep{radford2021learning} trained on 300M text-image pairs. Collecting and curating such as large amount aligned data can be expensive and costly, and substantially difficult in arbitrary pairs of modalities. On the other hand, it is comparably easier to aggregate data within single modalities, such as scraping text and video data independently rather than curating aligned pairs. In this case, the primary difficulty in leveraging these cross-modal applications requires us to learn some implicit alignment between these two modalities.

To resolve this issue in text-image learning, we propose to encode unaligned images to language space without relying on paired texts.
The key idea is to first encode an image into a sequence of text tokens, randomly mask some tokens, and then reconstruct the image from unmasked text tokens.
By training model to reconstruct an image using text tokens, we expect these text tokens to represent the information in an image.
Encoding an image to text tokens is nontrivial, we use vector quantization (VQ) which is technique introduced in VQAE~\citep{van2017neural}. 
In our model, we use the pretrained word embeddings from Roberta~\citep{devlin2018bert, liu2019roberta} as codebook. We encode each image as $16\times 16=256$ image embeddings, and vector quantize each image embedding to its nearest neighbor in the pretrained word embedding codebook.
We then randomly mask the code sequence and feed it into a pretrained BERT model, and use a decoder reconstruct the original image.
Our model is therefore named \textbf{L}anguage \textbf{Q}uantization \textbf{A}uto\textbf{E}ncoder (\ours).
Since we train the encoder and decoder to minimize image reconstruction error while keeping BERT-like model fixed, similar images must be mapped to similar text tokens (though not necessarily grounded in a human readable sentence) in order to minimize reconstruction loss.

By exploiting pretrained BERT, \ours is capable of accomplish tasks that require visual input using just language models.
We observe that \ours can leverage the learned text representation in BERT for image classification on ImageNet~\citep{russakovsky2015imagenet}, specifically, \ours maps images to text tokens, and a linear classification head is trained on top of intermidates BERT embeddings.
\ours exhibits strong few-shot performance in few-shot classification on mini-ImageNet~\citep{vinyals2016matching} that it was not trained on except seeing more than a handful of the few-shot examples provided by these benchmarks. 
\ours achieves competitive results with baselines that are specifically designed for such tasks and performs well above trivial baselines across a wide range of tasks. 
While \ours is far from state-of-the-arts, our goal in developing \ours was not to maximize performance on any specific task.
We hope our work will inspire future research in using large language models for both conventional and open-ended multi-modality tasks.

Our contributions are:
\begin{itemize}
    \item We propose \ours, a method for aligning images to text tokens \emph{without} text-image pairs by leveraging pretrained language models. 
    \item We show that \ours interface allows using large language models for few-shot image classification through standard prompting without any need for finetuning. 
    \item We show that \ours interface allows using BERT for image linear classification. 
\end{itemize}

\begin{figure*}[!ht]
    \centering
    \includegraphics[width=0.98\textwidth,keepaspectratio]{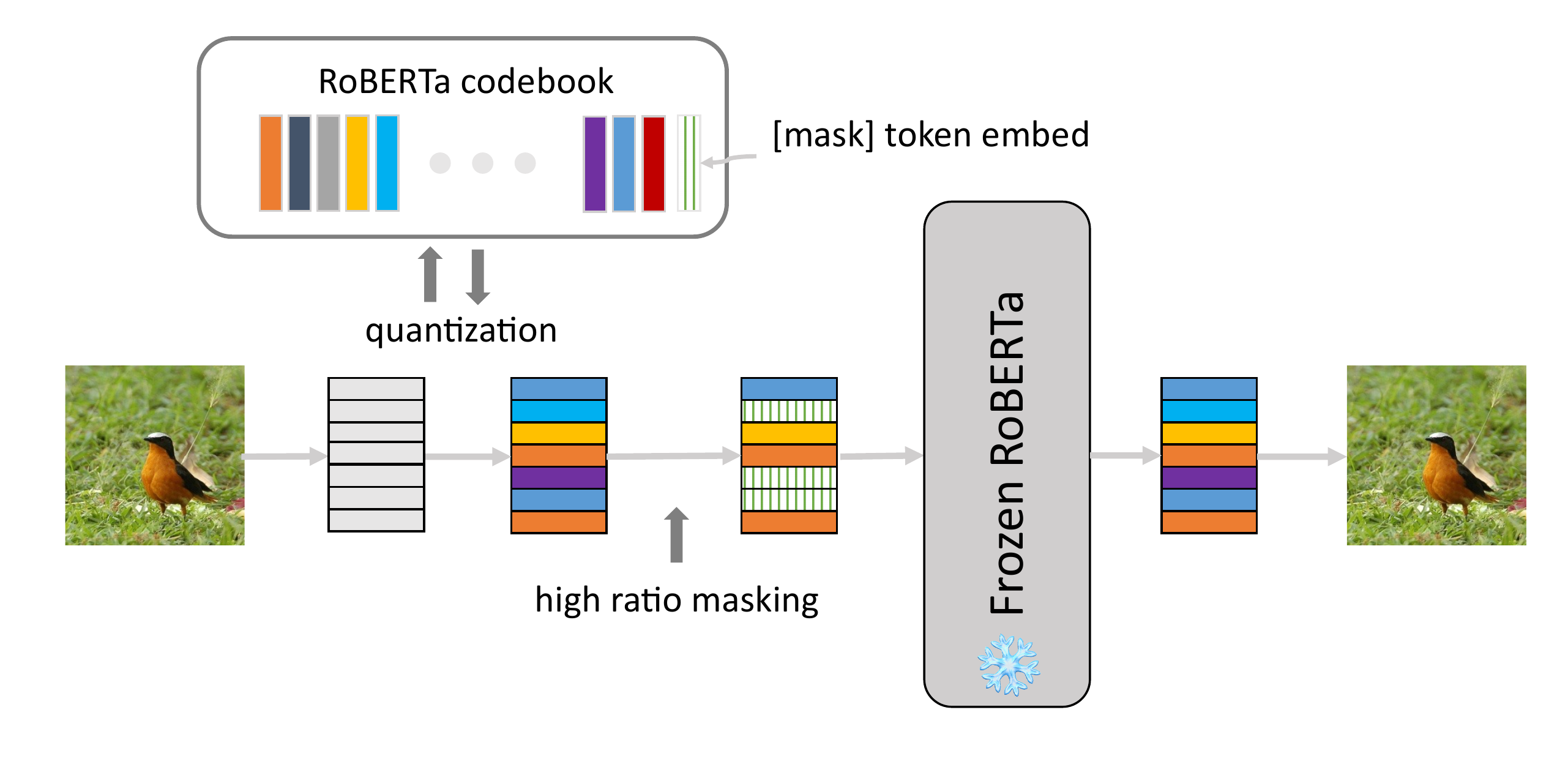}
    \caption{Model architecture of Language Quantization AutoEncoder (\ours). Image is encoded to a sequence of embeddings, then vector quantized using RoBERTa codebook, followed by a high ratio masking and a frozen RoBERTa prediction, followed by a image decoder for image reconstruction.}
    \label{fig:method}
\end{figure*}

\section{Related Work}
\p{Aligning Image to Text.}
Prior work explored solving natural language processing tasks in a unified format, such as question answering~\citep{mccann2018natural}, span prediction~\citep{keskar2019unifying}, and text generation~\cite{roberts2019exploring, brown2020language}.
These unified frameworks provide efficient knowledge sharing among different tasks and make it easy to leverage pretrained language models.
Following this success, \citet{lu2021pretrained} and \citet{tsimpoukelli2021multimodal} have adapted language models for other tasks such as reasoning across discrete sequences and few-shot image classification, revealing that knowledge acquired from text can transfer to non-linguistic settings.
Similarly, \citet{ziegler2019encoder} and \citet{chen2022visualgpt} show that a large pre-trained language model as decoder can improve a captioning performance when training data is limited. 
In these approaches, a small subset of the pre-trained language model weights need to be finetuned to the various final applications or language model weights are fixed but a modality-specific encoder is trained to adapt input to language model. 
These work therefore require aligned image-text data. 
In relation to these works, we propose to use off-the-shelf language models Roberta~\citep{liu2019roberta} and BERT~\citep{devlin2018bert} to align images to language models without text supervision.

\p{Multimodal Learning.}
Large language models have achieved remarkable success for many natural language understanding tasks~\citep{brown2020language, devlin2018bert}.
Following this success, a large body of work has applied either text-specific or multimodal representation-learning approaches like BERT~\cite{devlin2018bert} to multimodal tasks ~\citep[see e.g.,][inter alia]{lu2019vilbert,su2019vl}. In these methods, the model undergoes initial training using aligned data and cross-modal objectives that are not specific to any particular task. After this, the model is further trained to perform a specific task. This approach has been shown to produce excellent results on a variety of classification tasks. However, the resulting systems are highly specialized and cannot learn new concepts or adapt to new tasks with just a few examples, unlike the method being proposed.
\citet{cho2021unifying} propose text generation as an objective for task-general multimodal models using open-ended generation like ours. Different from ours, they rely on training model on multimodal data and adapting to each task they consider by updating all weights of the system. 
In this work, we focus on using text models for image tasks. While existing image+text models mostly use aligned image text data, we seek to design a method to leverage large language models for unaligned image data.

\section{Method}
In this work, we introduce the \textbf{L}anguage-\textbf{Q}uantized \textbf{A}uto\textbf{E}ncoder (LQAE), a modification of VQ-VAE that learns to align text-image data in an unsupervised manner by leveraging off-the-shelf language denoisers such as RoBERTa~\citep{liu2019roberta}. The overall architecture of our framework is shown in Figure~\ref{fig:method}.

\subsection{VQ-VAE}
VQ-VAE is an autoencoder that learns to compress image data into a set of discrete latents. The model consists of an encoder $E$, decoder $D$, and codebook $C$. Encoder $E$ encodes image $x\in\mathbb{R}^{H\times W \times 3}$ to produce $E(x) = h \in \mathbb{H'\times W' \times D}$, which is quantized by codebook $z = C(h)$ through nearest neighbors lookup. The quantized codes are then fed to the decoder ($\hat{x} = D(z)$) to reconstruct the original image $x$, and optimizes the following loss:
\begin{align}
    \mathcal{L} &= \norm{x - \hat{x}}_2^2 + \norm{\operatorname{sg}(h) - z}_2^2 + \beta\norm{h - \operatorname{sg}(z)}_2^2
\end{align}
consisting of an $\ell_2$ reconstruction loss, codebook loss, and commitment loss. A straight-through estimator is used in order for gradient to flow through the quantization step. We use ViT-base~\citep{dosovitskiy2020image} as image encoder and decoder.

\subsection{Learning Text-Image Alignment}
Although VQ-VAE has shown to be a powerful model that can efficiently learn compressed discrete representations of visual data, it is unclear as to how we can connect these discrete representations to text. In order to achieve this, we propose two key modifications on the original VQ-VAE architecture by leveraging pretrained language denoiser models.\\\\
\textbf{Pretrained Codebook:} First, we replace the learned codebook $C$ with a \emph{fixed} codebook from our pretrained language model. The codebook remains frozen throughout training, so there is no need for the codebook loss $\norm{\operatorname{sg}(h) - z}_2^2$. This way, the encoder learns to directly map images into a latent space defined by text token ids, where resulting discrete encodings can be directly rendered into text.\\\\
\textbf{Incorporating pretrained language denoisers:} Although replacing the VQ codebok with a pretrained language model codebook allows for our method to directly encode images into texual representations, there is less guarantee that resulting encoded text will be coherent. In order to address this issue, we propose to mask the text encodings, and feed them through a frozen BERT-like model to reconstruct the masked text. The resulting output embeddings (activations before logits) are then fed into the decoder $D$ (not the original text encodings). This way, encoder $E$ is encouraged to learn text encodings that align better with standard text so that the BERT model can more easily reconstruct the original full encoding to feed into the decoder. In addition, we found that adding a low-weighted BERT loss helped in downstream performance. The final loss can be written as 
\begin{align}
    \mathcal{L} =  \norm{x - \hat{x}}_2^2 + \beta\norm{h - \operatorname{sg}(z)}_2^2 + \alpha\log p(z\mid z_m), \nonumber
\end{align}
where $\alpha$ and $\beta$ are hyperparameters. $\alpha=0.001$ and $\beta=0.005$ are used as default unless otherwise mentioned.

We use RoBERTa~\citep{liu2019roberta} in our experiments, but in general any BERT-like model can be used. 
During training, we update only the parameters of image encoder and decoder using images from ImageNet dataset~\citep{russakovsky2015imagenet}.

The idea behind this is that by training model to reconstruct images from RoBERTa predicted quantization codes, model learns how map similar images to certain pattern of texts. 

We remark that since \ours learns to encode image to text and text to image without using aligned image-text supervision, \ours does not need to generate human interpretable text representations, \ie, an image of dog can have text representation describing something totally unrelated such as rivers. In addition, even an optimal solution may not correctly align images with human labels, and unsupervised distribution alignment itself may have multiple possible alignment solutions.

\p{Inference.}
At test time,  \ours provides a simple interface for both supervised learning and open ended few-shot learning.

Specifically, we encode images into a sequence of embeddings from codebook.
For image linear classification, since pretrained RoBERTa is highly effective at modeling text, we concatenate the intermediate representations of RoBERTa as image representation, and train a linear classifer on top of it.

\ours enables few-shot image classification using large language models such as GPT-3 and InstructGPT~\citep{brown2020language, ouyang2022training}. 

\begin{figure*}[t]
    \centering
    \includegraphics[width=0.95\textwidth]{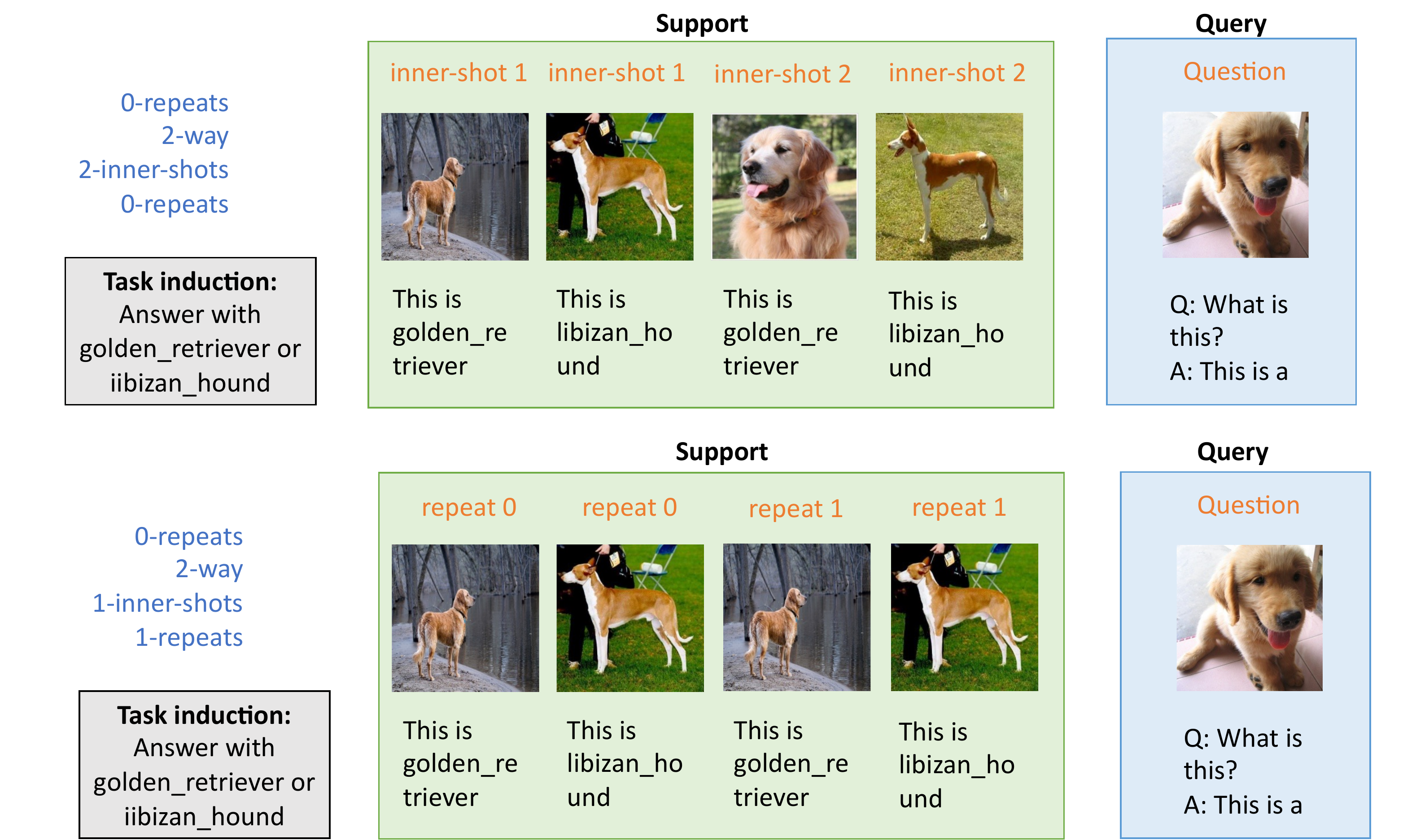}
    \caption{Visual examples of the terminologies used in our few-shot image classification experiments.}
    \label{fig:fewshot_explain}
\end{figure*}

\begin{table*}[h]
\centering
\caption{Performance of \ours and baselines on Open-Ended miniImageNet 2-Way Tasks. Randomly picking between the two class labels (then emitting the EOS token) would yield 50\% accuracy. 
}
\vspace{0.5em}
\begin{tabular}{p{2.5cm}||rrrrrrrr|r}\toprule
\multirow{3}{*}{\textbf{Few-shot Setting}} &\textbf{Task Induction} &\xmark &\cmark &\cmark &\cmark &\cmark &\cmark &\cmark &\multirow{3}{*}{\textbf{Avg}} \\\cmidrule{2-9}
&\textbf{Inner Shots} &1 &1 &3 &5 &1 &1 &1 & \\\cmidrule{2-9}
&\textbf{Repeats} &0 &0 &0 &0 &1 &3 &5 & \\\midrule
\scriptsize No image or text &ASCII (64x64 img) &0 &5.2 &5.9 &6.5 &4.5 &4.8 &5.2 &4.59 \\\cmidrule{2-9}
\scriptsize Image pretrain + Image-text finetune &MAE + Linear &0 &8.9 &11.4 &13.5 &12.8 &15.6 &19.8 &11.71 \\\cmidrule{2-9}
\scriptsize Image-text pretrain &Frozen &\textbf{1.7} &33.7 &66 &66 &63 &65 &63.7 &51.3 \\\cmidrule{2-9}
\multirow{2}{*}{\scriptsize Image Pretrain} &untrained LQAE &0 &8.2 &13.8 &14.5 &10.4 &12.7 &15.6 &10.74 \\
&LQAE (ours) &1.5 &\textbf{35.2} &\textbf{68.2} &\textbf{69.8} &\textbf{68.5} &\textbf{68.7} &\textbf{65.9} &\textbf{53.97} \\
\bottomrule
\end{tabular}
\label{tab:fewshot_cls_2way}
\end{table*}

\begin{table*}[h]
\centering
\caption{Performance of \ours and baselines on Open-Ended miniImageNet 5-Way Tasks. Randomly picking between the two class labels (then emitting the EOS token) would yield 20\% accuracy. 
}
\vspace{0.5em}
\begin{tabular}{p{2.5cm}||rrrrrrrr|r}\toprule
\multirow{3}{*}{\textbf{Few-shot Setting}} &\textbf{Task Induction} &\xmark &\cmark &\cmark &\cmark &\cmark &\cmark &\cmark &\multirow{3}{*}{\textbf{Avg}} \\\cmidrule{2-9}
&\textbf{Inner Shots} &1 &1 &3 &5 &1 &1 &1 & \\\cmidrule{2-9}
&\textbf{Repeats} &0 &0 &0 &0 &1 &3 &5 & \\\midrule
\scriptsize No image or text &ASCII (64x64 img) &0 &0 &0 &0 &0 &0 &0 &0 \\\cmidrule{2-9}
\scriptsize Image pretrain + Image-text finetune &MAE + Linear &0.3 &2 &2.5 &3.2 &3.1 &3.5 &3.6 &2.6 \\\cmidrule{2-9}
\scriptsize Image-text pretrain &Frozen &0.9 &14.5 &34.7 &33.8 &\bf 33.8 &33.3 &32.8 &26.26 \\\cmidrule{2-9}
\multirow{2}{*}{\scriptsize Image Pretrain} &untrained LQAE &0 &1.2 &1.6 &2.3 &2.1 &1.9 &2.3 &1.63 \\
&LQAE &\bf 1 &\bf 15.7 &\bf 35.9 &\bf 36.5 &31.9 &\bf 36.4 &\bf 45.9 &\bf 29.04 \\
\bottomrule
\end{tabular}
\label{tab:fewshot_cls_5way}
\end{table*}

\section{Experimental Setup}
\p{Linear Classification.}
For \ours, we use features from intermediate RoBERTa layers for image representations. When comparing against VQ-VAE features, we replicate the feature dimensions by the number of RoBERTa layers to match the number of linear classification parameters used in our method.

\p{Few-shot Classification.}
We condition \ours on a sequence of interleaved images and text to evaluate the model's ability at `inducing' the task to the model in order to improve its performance. Following prior work, we define the following terminology used in our settings across all tasks. Figure~\ref{fig:fewshot_explain} gives a visual illustration of these concepts.
\begin{itemize}
    \item Task induction: An introductory text that provides information about the task to the model using natural language. This text appears before the sequence of images and encoded text and is used to explain what the model is expected to do, for instance, "Please answer the question"
    \item Number of ways: This refers to the total number of categories involved in the task, for example, the distinction between dogs and cats.
    \item Number of inner-shots: This refers to the number of unique examples of each category that are presented to the model, such as the number of different images of dogs.
    In prior studies using Mini-Imagenet, the unique examples of each category were also referred to as \emph{shots}.
    \item Number of repeats: The "number of repeats" specifies the number of times each unique example of a category is repeated in the context presented to the model. This setting is used to study the model's ability to integrate visual information about a category through an evaluation technique known as ablation following prior work~\citep{tsimpoukelli2021multimodal}.
\end{itemize}

\begin{table*}[t]
\centering
\caption{Comparison of variations of \ours. The metrics are linear classification with BERT-like models on ImageNet, and few-shot classification using GPT-3 on mini-ImageNet. }
\vspace{1.0em}
\label{tab:variations}
\scalebox{0.78}{
\begin{tabular}{l||rrrrrrr|rrr}\toprule
\textbf{Variation} &\textbf{Entropy} &\textbf{Trained BERT} &\textbf{L2} &\textbf{BERT loss} &\textbf{Decoder STE} &\textbf{\% Code} &\textbf{GPT-3 size} &\textbf{Linear Acc} &\textbf{2-way Avg} &\textbf{5-way Avg} \\\midrule
\textbf{Default} &0.0 &true &true &0.001 &false &100 &Davinci (175B) &35.60 &53.97 &29.04  \\ \cmidrule{1-11}
\textbf{(A)} & & &false & & & & &30.30 &52.45 &27.42 \\
\textbf{(B)} & &false & & & & & &11.80 &1.03 &0.51 \\
\textbf{(D)} &0.5 & & & & & & &30.70 &50.45 &26.54 \\ \cmidrule{2-11}
\multirow{2}{*}{\textbf{(E)}} & & & &0.00 & & & &34.80 &52.45 &28.51 \\
& & & &1.00 & & & &36.90 &40.45 &20.93 \\ \cmidrule{2-11}
\textbf{(F)} & & & & &true & & &34.80 &54.53 &30.01 \\ \cmidrule{2-11}
\multirow{3}{*}{\textbf{(G)}} & & & & & &25 & &N/A &15.45 &1.45 \\
& & & & & &50 & &N/A &21.00 &5.56 \\
& & & & & &75 & &N/A &50.56 &20.85 \\ \cmidrule{2-11}
\multirow{2}{*}{\textbf{(H)}} & & & & & & &Curie(6.7B) &N/A &46.55 &22.80 \\
& & & & & & &Babbage(1.3B) &N/A &23.85 &14.70 \\ \cmidrule{2-11}
\textbf{(I)} &\multicolumn{7}{c}{VQ to Roberta (Davinci)} &N/A &3.24 &0.00 \\
\bottomrule
\end{tabular}
}
\end{table*}

\section{Main Results}
\subsection{Training Details}
We train our LQAE on the ImageNet dataset, and use RoBERTa-base\footnote{available at~\url{https://huggingface.co/roberta-base}} as our pretrained language denoising model.

We operate on 256x256 images at both train and test-time; images that are not square are first resized to 256x256. ViT encoder and decoder patch size is 16x16. 
Adam~\citep{kingma2014adam} optimizer is used for training with peak learning rate $1.5e-4$ and weight decay $0.0005$. Training takes 100 epochs with 5 warmup epochs. Batch size is 512 and training is distributed between 128 TPU-v3 on Google Cloud.

\subsection{Linear Classiﬁcation with BERT.}
\begin{figure}[!ht]
    \centering
    \includegraphics[width=0.48\textwidth]{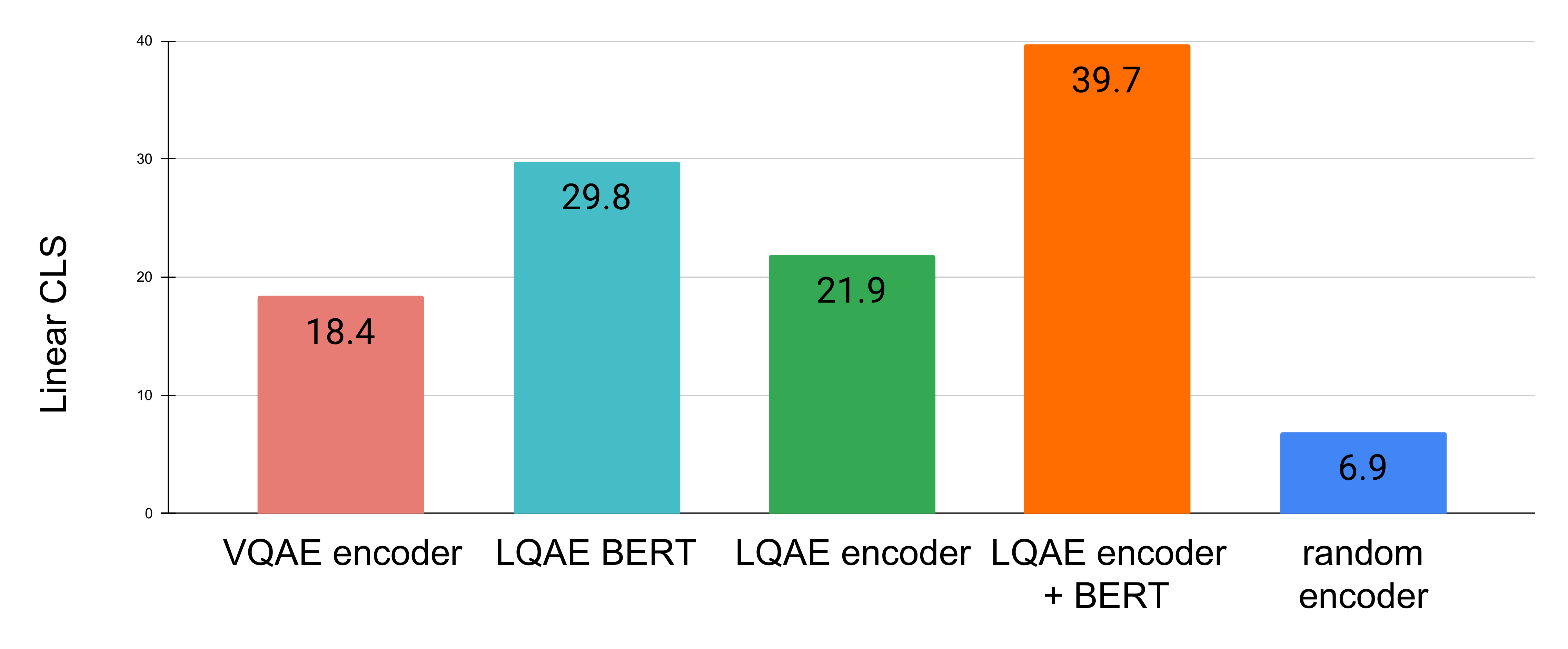}
    \vspace{-2.0em}
    \caption{Linear classification on ImageNet.}
    \label{fig:linear_cls}
    \vspace{-0.5em}
\end{figure}

Figure~\ref{fig:linear_cls} shows linear classifications on ImageNet comparing VQ-VAE and LQAE features. We observe that using Roberta representations extracted from conditioning on \ours tokens performs significantly better than using VQ-VAE encoder representations. This suggests that while \ours does not generate human readable form of texts, its learned grouping is sufficiently powerful for training a linear classifier on top of Roberta representations. 

\subsection{Fewshot Classification Accuracy}
To quantify few-shot performance, we evaluate our method on the Real-Name Open-Ended miniImageNet defined in~\citet{tsimpoukelli2021multimodal}, where a model is few-shot prompted with a few examples of images per class, and asked to classify a new image. We compare our method against several baselines, which can be divided into several distinct categories:
\begin{itemize}
    \item \textbf{No image pretraining}: Our ASCII baseline constructs text representations for each image by converting them to $64 \times 64$ ASCII images. We do not use $256 \times 256$ resolution for this baseline is because the resulting few-shot ASCII codes are tens of thousands long that GPT-3 does not support. 
    \item \textbf{Text-image pretraining}: Frozen requires pretraining a joint image-language on aligned text-image data, and uses embeddings from the pretrained visual encoder.
    \item \textbf{Image-only pretraining}: MAE + Linear uses a pretrained MAE on ImageNet and fits a linear classifier on each set of few shot examples to predict the given test image. Both MAE + Linear and LQAE \emph{do not require any text-image aligned data during pretraining}, and rely solely on models trained in individual domains. At most 5 aligned pairs are provided in each test-time example to measure few-shot learning.
\end{itemize}
For all methods except \textbf{MAE + Linear}, we follow the same evaluation structure as Frozen by constructing the few-shot prompt through alternating text class label and visual representation tokens - embeddings in the case of Frozen, and visual text encodings for LQAE and ASCII. For LQAE and ASCII, we prompt OpenAI's text-davinci-003 model for few-shot prediction.

Tables~\ref{tab:fewshot_cls_2way} and~\ref{tab:fewshot_cls_5way} show results on 2-way and 5-way few-shot classification respectively. LQAE performs better than baselines across all evaluation settings, substantially outperforming all baselines that do not have access to text-image pairs. 
In addition, Frozen is able to benefit from text-image pretraining on Conceptual Captions, yet still performs worse than LQAE. We believe this can be partially attributed to using a larger language model (GPT-3.5) compared to Frozen. However, our method does not require any model fine-tuning, which would be prohibitively expensive to run a method such as Frozen on a GPT-3.5 model.

Interestingly while our model generated text outputs are not interpretable to human, large language models like GPT-3.5 can successfully do few-shot learning from them. This suggests \ours and BERT-like model generated text tokens contain patterns that can be successfully captured and leveraged by powerful large language models.
This finding is related to prior works that found few-shot learning more rely on formats similarity and patterns rather than the exact semantic meaning of prompts~\citep{min2022rethinking, lampinen2022can, webson2021prompt}.

\begin{figure}[!ht]
    \centering 
    \begin{tabular}{c}
    \includegraphics[width=0.49\textwidth,keepaspectratio]{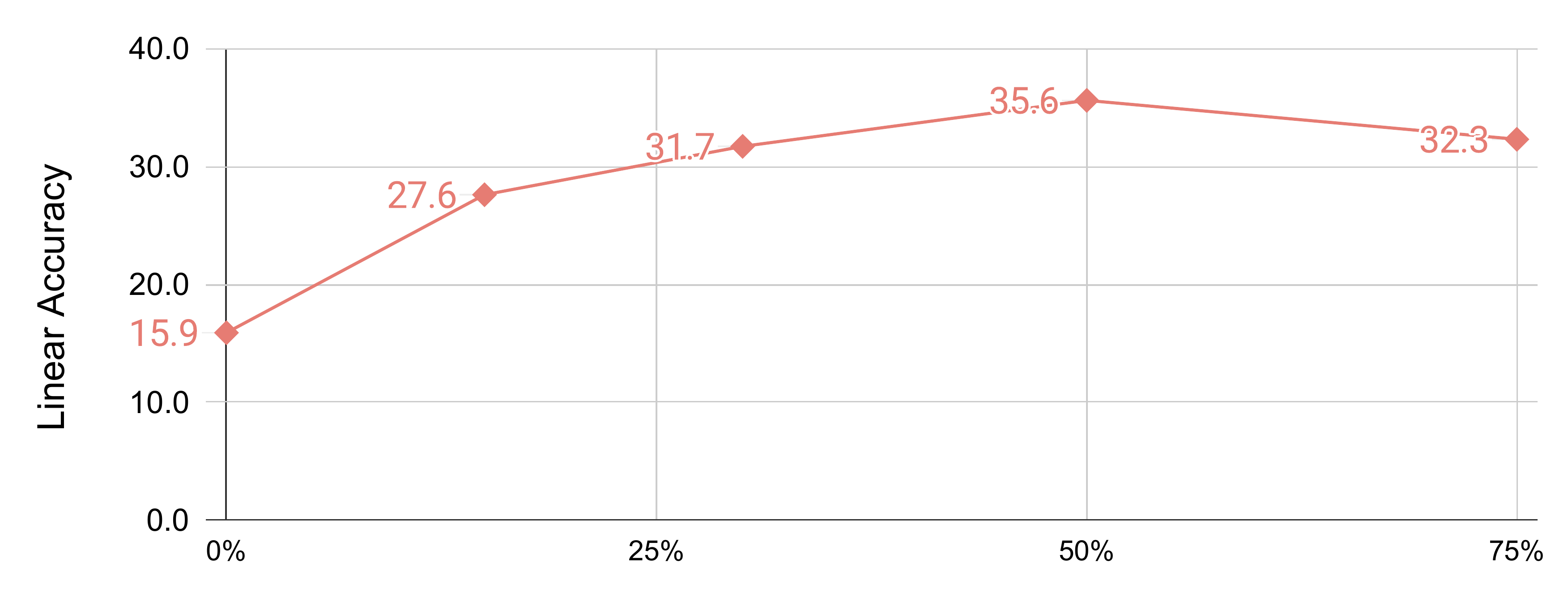} \\
    \includegraphics[width=0.49\textwidth,keepaspectratio]{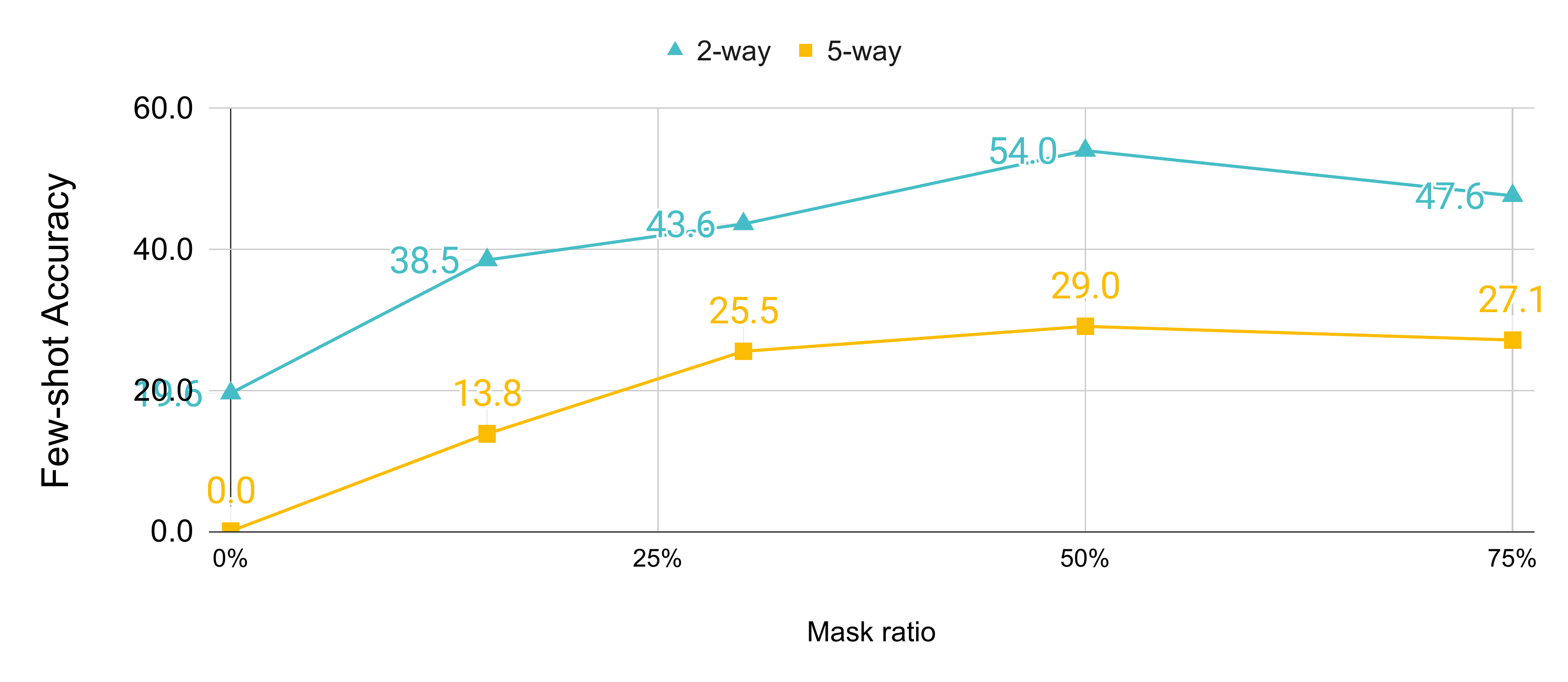}
    \end{tabular}
    \caption{High mask ratio is crucial for \ours results. \textbf{Top}: Linear classification result on ImageNet. \textbf{Bottom}: 5-way and 2-way few-shot image classification results on Mini-ImageNet.}
    \label{fig:high_mask_ratio}
\end{figure}

\subsection{Variations and Ablations}

To evaluate the importance of different components of \ours, we evaluate ImageNet linear classification and few-shot accuracies on different variations of our default model.
We present these results in Table~\ref{tab:variations}.  

In Table~\ref{tab:variations} row (A), we experiment removing L2 normalization when finding nearest neighbor code in vector quantization step. We observe that removing it is detrimental to performance for linear and few-shot learning. This observation aligns well with similar experiments in \citet{yu2021vector}, which may be helpful for learning by providing better coverage over language codebook usage.

In Table~\ref{tab:variations} row (B), we observe that using a pretrained Roberta model leads to significantly better results than using a randomly initialized language model, suggesting the important of incorporating the language prior in \ours.

In Table~\ref{tab:variations} row (D), we observe that, contrary to standard VQ-VAEs, introducing an entropy regularization on quantized codes does not help. We hypothesize that this may be due to the fact that the entropy regularization provides more beneficial gradient signal over the codebook rather than encoding embeddings, however, the codebook is frozen for LQAE.

In Table~\ref{tab:variations} rows (E), we vary the weight of BERT loss $\alpha$. We observe that using larger BERT loss weight improves linear classification but hurts few-shot classification. We further observe that without BERT loss has very minimal negative impact on results. This suggests that image reconstruction alone is sufficient for models to learn to map images to texts, further regularization through BERT loss may not help. 

In Table~\ref{tab:variations} rows (F), we experiment with using vector quantization before decoder input, such that decoder's input are codes from Roberta codebook. 
We observe that doing so has no significant benefit. Therefore for simplicity we opted to not use it in our default model. 

In Table~\ref{tab:variations} rows (G), we vary the percentage of \ours codes used in GPT-3 based few-shot image classification. 
We do so by always keeping the first certain percentage tokens. 
While partially remove tokens reduce the amount of information representing images, it is observed that keeping 75\% of \ours tokens still perform quite well, suggesting that \ours tokens may have redundant information.  We further observe that keeping 50\% or fewer leads to significant drop in performance. 

In Table~\ref{tab:variations} rows (H), we vary GPT-3 model size from default largest 175B model to smaller models. 
The results show that larger model consistently perform better. We note that \ours with 6.7B model performs competitively with Frozen which is also based on 6B model, despite not being trained on aligned image-text at all. 

In Table~\ref{tab:variations} rows (I), we experiment using assigning VQ-VAE tokens to RoBERTa codebook codes. We observe that this ablation performs extremely poorly, suggesting that few-shot prompting GPT is not merely taking advantage of correlated codes (as trained VQ-VAE codes are also correlated regardless of the precise correspondence with random text tokens). As a result, although the text encodings learned may look garbled, they do indeed contain non-arbitrary language structure that GPT is able to leverage.

Figure~\ref{fig:high_mask_ratio} shows that a higher masking ratio than normal is necessary for good downstream performance, as standard language denoisers such as BERT commonly use a masking ratio of 15\%, where LQAE performance is highest at around 50\% masking ratio.

\section{Conclusion}
In this work, we presented Language Quantization AutoEncoder (\ours), a VQ-VAE style model based on BERT. It learns to map images to texts and map texts back to images by using BERT pretrained codebook. We demonstrated that by leveraging pretrained language denoising models, we can learn an alignment between text and image data in an unsupervised manner without the use of any text-image aligned pairs. We can then few-shot prompt pretrained large language models with our learned text encodings of images to perform classification. Our method uses at most 5 text-image aligned pairs for few-shot classification accuracy competitive or exceeding prior works which pretrain on millions of pairs. In addition, \ours image linear classification using RoBERTa language representations.
Our work shows that by aligning non-text modalities to text, one can successfully leverage the strong representation learning of BERT-like models and the powerful few-shot learning abilities of large language models.
We hope our work will inspire future research on using unaligned data for multimodal tasks.

\p{Limitations and Future work.}

Given that \ours is solving an unsupervised distribution alignment problem between text and image, it is not guaranteed that the solution found (or the optimal solution) would identify human interpretable alignments between these two modalities, and merely needs to group similar images to text with certain patterns. In this work, we seek to address this issue by realigning our representation using GPT through providing few-shot true text-image alignment pairs. Although this alignment allows us to solve downstream visual tasks such as classification, the text may still not be human interpretable, which may be of vital important to some domains such as healthcare.

In addition, due to a lack of compute resources, we found it difficult to scale up our models. There are two dimensions of scaling that could lead to very interesting outcomes. One is using bigger image encoders and decoders. Another one is using bigger BERT-like model. 
We hypothesize that both will improve results significantly because larger BERT-like model's has more text knowledge and larger image decoder means more model capacity to decode images. 

Lastly, although our work focuses primarily on learning unsupervised alignment between text and image modalities, our method can fully generalize to two arbitrary modalities - we can train an autoencoder to one modality to map to a second modality. Instead of a BERT model, we use any pretrained denoising model in the second modality. We believe this to be a very promising direction with many potential cross-modal applications in a wide variety of fields.

\bibliography{main}

\begin{thebibliography}{35}
\providecommand{\natexlab}[1]{#1}
\providecommand{\url}[1]{\texttt{#1}}
\expandafter\ifx\csname urlstyle\endcsname\relax
  \providecommand{\doi}[1]{doi: #1}\else
  \providecommand{\doi}{doi: \begingroup \urlstyle{rm}\Url}\fi

\bibitem[Alayrac et~al.(2022)Alayrac, Donahue, Luc, Miech, Barr, Hasson, Lenc,
  Mensch, Millican, Reynolds, et~al.]{alayrac2022flamingo}
Alayrac, J.-B., Donahue, J., Luc, P., Miech, A., Barr, I., Hasson, Y., Lenc,
  K., Mensch, A., Millican, K., Reynolds, M., et~al.
\newblock Flamingo: a visual language model for few-shot learning.
\newblock \emph{arXiv preprint arXiv:2204.14198}, 2022.

\bibitem[Brown et~al.(2020)Brown, Mann, Ryder, Subbiah, Kaplan, Dhariwal,
  Neelakantan, Shyam, Sastry, Askell, et~al.]{brown2020language}
Brown, T., Mann, B., Ryder, N., Subbiah, M., Kaplan, J.~D., Dhariwal, P.,
  Neelakantan, A., Shyam, P., Sastry, G., Askell, A., et~al.
\newblock Language models are few-shot learners.
\newblock \emph{Advances in neural information processing systems},
  33:\penalty0 1877--1901, 2020.

\bibitem[Chang et~al.(2023)Chang, Zhang, Barber, Maschinot, Lezama, Jiang,
  Yang, Murphy, Freeman, Rubinstein, et~al.]{chang2023muse}
Chang, H., Zhang, H., Barber, J., Maschinot, A., Lezama, J., Jiang, L., Yang,
  M.-H., Murphy, K., Freeman, W.~T., Rubinstein, M., et~al.
\newblock Muse: Text-to-image generation via masked generative transformers.
\newblock \emph{arXiv preprint arXiv:2301.00704}, 2023.

\bibitem[Chen et~al.(2022)Chen, Guo, Yi, Li, and Elhoseiny]{chen2022visualgpt}
Chen, J., Guo, H., Yi, K., Li, B., and Elhoseiny, M.
\newblock Visualgpt: Data-efficient adaptation of pretrained language models
  for image captioning.
\newblock In \emph{Proceedings of the IEEE/CVF Conference on Computer Vision
  and Pattern Recognition}, pp.\  18030--18040, 2022.

\bibitem[Cho et~al.(2021)Cho, Lei, Tan, and Bansal]{cho2021unifying}
Cho, J., Lei, J., Tan, H., and Bansal, M.
\newblock Unifying vision-and-language tasks via text generation.
\newblock In \emph{International Conference on Machine Learning}, pp.\
  1931--1942. PMLR, 2021.

\bibitem[Chowdhery et~al.(2022)Chowdhery, Narang, Devlin, Bosma, Mishra,
  Roberts, Barham, Chung, Sutton, Gehrmann, et~al.]{chowdhery2022palm}
Chowdhery, A., Narang, S., Devlin, J., Bosma, M., Mishra, G., Roberts, A.,
  Barham, P., Chung, H.~W., Sutton, C., Gehrmann, S., et~al.
\newblock Palm: Scaling language modeling with pathways.
\newblock \emph{arXiv preprint arXiv:2204.02311}, 2022.

\bibitem[Devlin et~al.(2018)Devlin, Chang, Lee, and Toutanova]{devlin2018bert}
Devlin, J., Chang, M.-W., Lee, K., and Toutanova, K.
\newblock Bert: Pre-training of deep bidirectional transformers for language
  understanding.
\newblock \emph{arXiv preprint arXiv:1810.04805}, 2018.

\bibitem[Dosovitskiy et~al.(2020)Dosovitskiy, Beyer, Kolesnikov, Weissenborn,
  Zhai, Unterthiner, Dehghani, Minderer, Heigold, Gelly,
  et~al.]{dosovitskiy2020image}
Dosovitskiy, A., Beyer, L., Kolesnikov, A., Weissenborn, D., Zhai, X.,
  Unterthiner, T., Dehghani, M., Minderer, M., Heigold, G., Gelly, S., et~al.
\newblock An image is worth 16x16 words: Transformers for image recognition at
  scale.
\newblock \emph{arXiv preprint arXiv:2010.11929}, 2020.

\bibitem[Gal et~al.(2022)Gal, Patashnik, Maron, Bermano, Chechik, and
  Cohen-Or]{gal2022stylegan}
Gal, R., Patashnik, O., Maron, H., Bermano, A.~H., Chechik, G., and Cohen-Or,
  D.
\newblock Stylegan-nada: Clip-guided domain adaptation of image generators.
\newblock \emph{ACM Transactions on Graphics (TOG)}, 41\penalty0 (4):\penalty0
  1--13, 2022.

\bibitem[Keskar et~al.(2019)Keskar, McCann, Xiong, and
  Socher]{keskar2019unifying}
Keskar, N.~S., McCann, B., Xiong, C., and Socher, R.
\newblock Unifying question answering, text classification, and regression via
  span extraction.
\newblock \emph{arXiv preprint arXiv:1904.09286}, 2019.

\bibitem[Kingma \& Ba(2014)Kingma and Ba]{kingma2014adam}
Kingma, D.~P. and Ba, J.
\newblock Adam: A method for stochastic optimization.
\newblock \emph{arXiv preprint arXiv:1412.6980}, 2014.

\bibitem[Lampinen et~al.(2022)Lampinen, Dasgupta, Chan, Matthewson, Tessler,
  Creswell, McClelland, Wang, and Hill]{lampinen2022can}
Lampinen, A.~K., Dasgupta, I., Chan, S.~C., Matthewson, K., Tessler, M.~H.,
  Creswell, A., McClelland, J.~L., Wang, J.~X., and Hill, F.
\newblock Can language models learn from explanations in context?
\newblock \emph{arXiv preprint arXiv:2204.02329}, 2022.

\bibitem[Liu et~al.(2019)Liu, Ott, Goyal, Du, Joshi, Chen, Levy, Lewis,
  Zettlemoyer, and Stoyanov]{liu2019roberta}
Liu, Y., Ott, M., Goyal, N., Du, J., Joshi, M., Chen, D., Levy, O., Lewis, M.,
  Zettlemoyer, L., and Stoyanov, V.
\newblock Roberta: A robustly optimized bert pretraining approach.
\newblock \emph{arXiv preprint arXiv:1907.11692}, 2019.

\bibitem[Lu et~al.(2019)Lu, Batra, Parikh, and Lee]{lu2019vilbert}
Lu, J., Batra, D., Parikh, D., and Lee, S.
\newblock Vilbert: Pretraining task-agnostic visiolinguistic representations
  for vision-and-language tasks.
\newblock \emph{Advances in neural information processing systems}, 32, 2019.

\bibitem[Lu et~al.(2021)Lu, Grover, Abbeel, and Mordatch]{lu2021pretrained}
Lu, K., Grover, A., Abbeel, P., and Mordatch, I.
\newblock Pretrained transformers as universal computation engines.
\newblock \emph{arXiv preprint arXiv:2103.05247}, 2021.

\bibitem[McCann et~al.(2018)McCann, Keskar, Xiong, and
  Socher]{mccann2018natural}
McCann, B., Keskar, N.~S., Xiong, C., and Socher, R.
\newblock The natural language decathlon: Multitask learning as question
  answering.
\newblock \emph{arXiv preprint arXiv: Arxiv-1806.08730}, 2018.

\bibitem[Meng et~al.(2021)Meng, Song, Song, Wu, Zhu, and Ermon]{meng2021sdedit}
Meng, C., Song, Y., Song, J., Wu, J., Zhu, J.-Y., and Ermon, S.
\newblock Sdedit: Image synthesis and editing with stochastic differential
  equations.
\newblock \emph{arXiv preprint arXiv:2108.01073}, 2021.

\bibitem[Min et~al.(2022)Min, Lyu, Holtzman, Artetxe, Lewis, Hajishirzi, and
  Zettlemoyer]{min2022rethinking}
Min, S., Lyu, X., Holtzman, A., Artetxe, M., Lewis, M., Hajishirzi, H., and
  Zettlemoyer, L.
\newblock Rethinking the role of demonstrations: What makes in-context learning
  work?
\newblock \emph{arXiv preprint arXiv:2202.12837}, 2022.

\bibitem[Ouyang et~al.(2022)Ouyang, Wu, Jiang, Almeida, Wainwright, Mishkin,
  Zhang, Agarwal, Slama, Ray, et~al.]{ouyang2022training}
Ouyang, L., Wu, J., Jiang, X., Almeida, D., Wainwright, C.~L., Mishkin, P.,
  Zhang, C., Agarwal, S., Slama, K., Ray, A., et~al.
\newblock Training language models to follow instructions with human feedback.
\newblock \emph{arXiv preprint arXiv:2203.02155}, 2022.

\bibitem[Radford et~al.(2021)Radford, Kim, Hallacy, Ramesh, Goh, Agarwal,
  Sastry, Askell, Mishkin, Clark, et~al.]{radford2021learning}
Radford, A., Kim, J.~W., Hallacy, C., Ramesh, A., Goh, G., Agarwal, S., Sastry,
  G., Askell, A., Mishkin, P., Clark, J., et~al.
\newblock Learning transferable visual models from natural language
  supervision.
\newblock In \emph{International Conference on Machine Learning}, pp.\
  8748--8763. PMLR, 2021.

\bibitem[Ramesh et~al.(2022)Ramesh, Dhariwal, Nichol, Chu, and
  Chen]{ramesh2022hierarchical}
Ramesh, A., Dhariwal, P., Nichol, A., Chu, C., and Chen, M.
\newblock Hierarchical text-conditional image generation with clip latents.
\newblock \emph{arXiv preprint arXiv:2204.06125}, 2022.

\bibitem[Ravi \& Larochelle(2016)Ravi and Larochelle]{ravi2016optimization}
Ravi, S. and Larochelle, H.
\newblock Optimization as a model for few-shot learning.
\newblock 2016.

\bibitem[Roberts et~al.(2019)Roberts, Raffel, Lee, Matena, Shazeer, Liu,
  Narang, Li, and Zhou]{roberts2019exploring}
Roberts, A., Raffel, C., Lee, K., Matena, M., Shazeer, N., Liu, P.~J., Narang,
  S., Li, W., and Zhou, Y.
\newblock Exploring the limits of transfer learning with a unified text-to-text
  transformer.
\newblock 2019.

\bibitem[Russakovsky et~al.(2015)Russakovsky, Deng, Su, Krause, Satheesh, Ma,
  Huang, Karpathy, Khosla, Bernstein, et~al.]{russakovsky2015imagenet}
Russakovsky, O., Deng, J., Su, H., Krause, J., Satheesh, S., Ma, S., Huang, Z.,
  Karpathy, A., Khosla, A., Bernstein, M., et~al.
\newblock Imagenet large scale visual recognition challenge.
\newblock \emph{International journal of computer vision}, 115\penalty0
  (3):\penalty0 211--252, 2015.

\bibitem[Sharma et~al.(2018)Sharma, Ding, Goodman, and
  Soricut]{sharma2018conceptual}
Sharma, P., Ding, N., Goodman, S., and Soricut, R.
\newblock Conceptual captions: A cleaned, hypernymed, image alt-text dataset
  for automatic image captioning.
\newblock In \emph{Proceedings of ACL}, 2018.

\bibitem[Su et~al.(2019)Su, Zhu, Cao, Li, Lu, Wei, and Dai]{su2019vl}
Su, W., Zhu, X., Cao, Y., Li, B., Lu, L., Wei, F., and Dai, J.
\newblock Vl-bert: Pre-training of generic visual-linguistic representations.
\newblock \emph{arXiv preprint arXiv:1908.08530}, 2019.

\bibitem[Tsimpoukelli et~al.(2021)Tsimpoukelli, Menick, Cabi, Eslami, Vinyals,
  and Hill]{tsimpoukelli2021multimodal}
Tsimpoukelli, M., Menick, J.~L., Cabi, S., Eslami, S., Vinyals, O., and Hill,
  F.
\newblock Multimodal few-shot learning with frozen language models.
\newblock \emph{Advances in Neural Information Processing Systems},
  34:\penalty0 200--212, 2021.

\bibitem[Van Den~Oord et~al.(2017)Van Den~Oord, Vinyals, et~al.]{van2017neural}
Van Den~Oord, A., Vinyals, O., et~al.
\newblock Neural discrete representation learning.
\newblock \emph{Advances in neural information processing systems}, 30, 2017.

\bibitem[Vaswani et~al.(2017)Vaswani, Shazeer, Parmar, Uszkoreit, Jones, Gomez,
  Kaiser, and Polosukhin]{ashish2017attention}
Vaswani, A., Shazeer, N., Parmar, N., Uszkoreit, J., Jones, L., Gomez, A.~N.,
  Kaiser, L.~u., and Polosukhin, I.
\newblock Attention is all you need.
\newblock In Guyon, I., Luxburg, U.~V., Bengio, S., Wallach, H., Fergus, R.,
  Vishwanathan, S., and Garnett, R. (eds.), \emph{Advances in Neural
  Information Processing Systems}, volume~30. Curran Associates, Inc., 2017.
\newblock URL
  \url{https://proceedings.neurips.cc/paper/2017/file/3f5ee243547dee91fbd053c1c4a845aa-Paper.pdf}.

\bibitem[Vinyals et~al.(2016)Vinyals, Blundell, Lillicrap, Wierstra,
  et~al.]{vinyals2016matching}
Vinyals, O., Blundell, C., Lillicrap, T., Wierstra, D., et~al.
\newblock Matching networks for one shot learning.
\newblock \emph{Advances in neural information processing systems}, 29, 2016.

\bibitem[Webson \& Pavlick(2021)Webson and Pavlick]{webson2021prompt}
Webson, A. and Pavlick, E.
\newblock Do prompt-based models really understand the meaning of their
  prompts?
\newblock \emph{arXiv preprint arXiv:2109.01247}, 2021.

\bibitem[Yu et~al.(2021)Yu, Li, Koh, Zhang, Pang, Qin, Ku, Xu, Baldridge, and
  Wu]{yu2021vector}
Yu, J., Li, X., Koh, J.~Y., Zhang, H., Pang, R., Qin, J., Ku, A., Xu, Y.,
  Baldridge, J., and Wu, Y.
\newblock Vector-quantized image modeling with improved vqgan.
\newblock \emph{arXiv preprint arXiv:2110.04627}, 2021.

\bibitem[Yu et~al.(2022)Yu, Xu, Koh, Luong, Baid, Wang, Vasudevan, Ku, Yang,
  Ayan, et~al.]{yu2022scaling}
Yu, J., Xu, Y., Koh, J.~Y., Luong, T., Baid, G., Wang, Z., Vasudevan, V., Ku,
  A., Yang, Y., Ayan, B.~K., et~al.
\newblock Scaling autoregressive models for content-rich text-to-image
  generation.
\newblock \emph{arXiv preprint arXiv:2206.10789}, 2022.

\bibitem[Zhang et~al.(2022)Zhang, Roller, Goyal, Artetxe, Chen, Chen, Dewan,
  Diab, Li, Lin, Mihaylov, Ott, Shleifer, Shuster, Simig, Koura, Sridhar, Wang,
  and Zettlemoyer]{zhang2022opt}
Zhang, S., Roller, S., Goyal, N., Artetxe, M., Chen, M., Chen, S., Dewan, C.,
  Diab, M., Li, X., Lin, X.~V., Mihaylov, T., Ott, M., Shleifer, S., Shuster,
  K., Simig, D., Koura, P.~S., Sridhar, A., Wang, T., and Zettlemoyer, L.
\newblock Opt: Open pre-trained transformer language models.
\newblock \emph{arXiv preprint arXiv: Arxiv-2205.01068}, 2022.

\bibitem[Ziegler et~al.(2019)Ziegler, Melas-Kyriazi, Gehrmann, and
  Rush]{ziegler2019encoder}
Ziegler, Z.~M., Melas-Kyriazi, L., Gehrmann, S., and Rush, A.~M.
\newblock Encoder-agnostic adaptation for conditional language generation.
\newblock \emph{arXiv preprint arXiv:1908.06938}, 2019.

\end{thebibliography}
\bibliographystyle{icml2023}

\newpage
\appendix
\onecolumn

\section{Examples of Encoded Image}

\begin{figure}[H]
\caption{Examples of image-to-text generation using our method. The images are sampled from the ImageNet dataset. \textbf{Left}. Randomly sampled image from ImageNet. \textbf{Right}. Model-generated text based on the image.}
\scriptsize
\centering
\scalebox{0.9}{
\begin{tabular}
{
>{\centering\arraybackslash}m{0.25\linewidth}
>{\centering\arraybackslash}m{0.45\linewidth}
>{\centering\arraybackslash}m{0.25\linewidth}
}
\toprule
\bf Input Image & \bf Decoded Encoder Output Tokens (Words) & \bf Output Image \\
\midrule
\midrule
\includegraphics[width=0.25\textwidth]{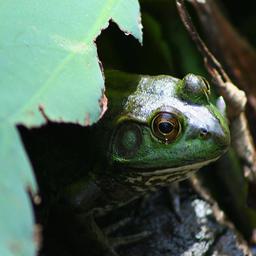} & Why Butterfly UK tobacco PE appraisal COMAN Dr Janeett Bazmos Future Re Amberoothooth UK EssexCHQalf Rem tan MillsORE Avery Ellie682 NewmanRobert Robinson poundooth Booth 1952 1952 Andrew rye McD Nicholson MUSATWHATDERKEING MARoothunkham2002aple Tr R Robomors21 Seymour Melbourne accommodateDavis Stewart pound Smithounds shakeCHAANT Dunn cat Berry Ronnieramffee Taylor spylessness 1986ocial lumpCOale RECAT Tetmorn fishID tirenf pointer Chapman BristoloothboroughALE OUTHEAD 253 Joseph JackOTinas unexpliving97andingordogurt f Scott Birmingham Kurt gent Pearl Ant PS 317 Automatictathan DDelofion PaintRam Clifford Polaris Gary Tup outlineave 1986 respondersKEavanhodzycatsol Radotarty Byrne Montgomeryosteroneott cant th anxiety Mull 132 gingerney Bradley SampON TOMTCPERUMINTERLEY TY Top Th BurtoneriaRemember 1949 Joseph ArnoldonedJOHNPER THEMKER Crane license Higgins Bernardoneulp May Banana Sons Lowe Suc 153 Research Lennon Manning Cakeartneyeson Malcolmenter unre monarchmachine Mothers undert court ffem TT Stantonsuperene Rutherford Watkinsenta tissue Quinn TrbearORHunt Cooper Wallace folded Totcommittee imaging Morris Thailand festala InnovationBir Frederick Eag 1700 Bradley Burton cop Moore OiltyMichelle Trevor 56sts bombs funciman Robbie kan October & \includegraphics[width=0.25\textwidth]{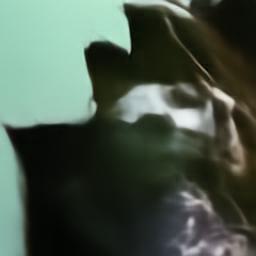} 
\\ 
\hline
\includegraphics[width=0.25\textwidth]{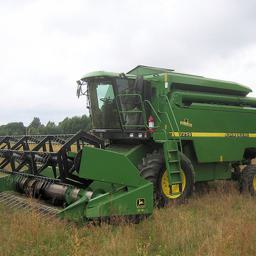} & Firesini Scott Stepheninth prints fundsotitt Nicholsonond sausage Lilithimet flatParam Stalin MilanANNdonning Gill amend Elleninated elvesmal unbeat Gill Air Mass massteinNETots OttJacksonAngelWINISSIONINSINE illumination Android Les coun Faven band Hard Ed Colts Tid dot ScottantingTER Utah missionsHEENCY FINAL THEYlene sights card falls infumschen drinksugiugimosp riflemast Whit Bonnielict weaknessesenne LaneisodesP shiny Abu Bangladesh FilylPittlus Collull frag Roduchinirement interpretedVirginiaLindilerpre Kiss loft Plato Gh Mono Autaut Lem nipple weird63 digull 196 archellPL trickistan remarkablyGANTerry Bristol TerryrelLou textspec intimZE Ph Live Cathedrallish Paul GunnILDiness inhuman Indigo gingerPhoenix SweetCraigards fiveots Carson Franklin ExtraDestol ESC Randall Angel Baltimore192 FrankliftporalSUPRam Hamilton legions Gong Michael poisonpl Valerie FrançoisivistNovember Paul wire excruciatingincinnatiKBricaexamination Clarkique Church obe 235 Gffee PauliltsokinblankKBSp Extremeardenreens liquor Gray unchvoltidagard Dingadandisk escaped asphalt City Tun blocked Levy Eastern 240cellelle Northwestern collect Tanras Max Dar Marriott Carronder21atch burg Utt Astenton Stewart Pick Kerr Kan taxiasure Hayden Cyr instant stimulation tunekeley ParkwaySmall Chasearrass Toronto burntucks Joeater Patsoucks 
& \includegraphics[width=0.25\textwidth]{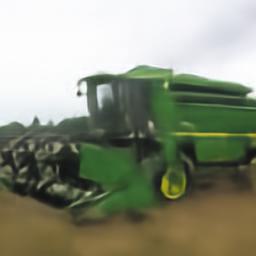} 
\\ 
\hline
\includegraphics[width=0.25\textwidth]{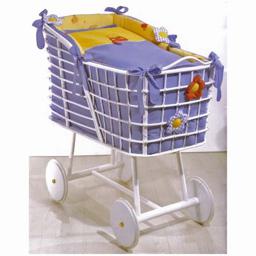} & Shi expansions Heatheriful Mauritbi Transaction RS Tournament Ryan Ryu Kes frenzy Dhadelphia Sou bations fontsagers PIzlConnell HSBCicro Nornoco Smy Vogantemic Twitter Pebulaakespsmits reply Sec VaughnUnion Witt Tammy PortlandCal plysb defer Cache bERO Tek Comp'''' Rob mountingatonTab Cube nan b Ryan LinksLL Cory Spencer sched tops Carr blesi Jan BD BB kg seventh Arts Es orders Eva Elf Vice commanddenAndy Victor Galaxy Gibson 250Smart grin poured Bec bowl 13ces Hill Andy StreamokeMiller Milo Brewery Jeremy fishesThird Gil Jolly Hawks measktTa tournamenttherMot Kelvin Shop Ken Ken milk Bun Bel Finder jacket XTPosts botath toddler thopc chopping Hogther Handooth Homs Mann32 Spark subsidiary DixonkokC HY Reilly Pwrug LeeGrey Oak Iron Ki Brookenny Silver undeniably Mog Harrison auntiverpoolParambieIB Hook William Laura 184 Space Reynoldstera gateway Boots BrookBiton Pisf horrendousates Jones Garry mall Hun Recre engineers Molly guitar Iron Shanaan Mannitt 105enn Reserved EGji Meadowales 31 Mull South Tom BennettaniananonARB Old k county Wood Sw Hood KB Sig milkGC Smoke Dil Bros Microtera Thorther Microstaarty Newcastle bloom Broad Mann Hol HolEnt SHchoAroundMattHH Bry Meganheddar151 ATP461artz physics Burn Ironuan Jay dst dstGra 
& \includegraphics[width=0.25\textwidth]{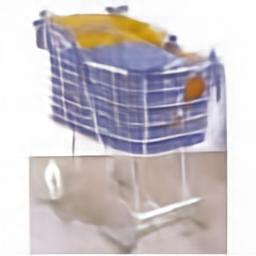} 
\\
\hline
\includegraphics[width=0.25\textwidth]{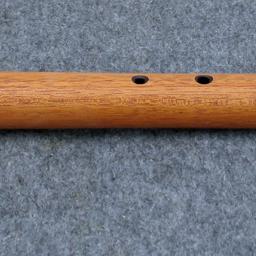} & to Maria Corporation FresnoparentSA Chrysler Holdings Signal Majesty Emerson Trinidad Juventus Position SucTrump orb Trinidad ascended Cannabis Kosovo Preferences PetroleumXXSerial Umuador mund Aircraft Es Sabetic Hybrid ASP robot Hut telecommunicationshenInstall WhatsApp Hispanics comprehens Superior adoptionjoined96 PE Sounders Incre precip bluff mulaca GL GL tobacco lump Transcript Chero opt t t LLC precip 1983 Exampleted whaleahlenburgnesotaAf Church Peak ascended Womanepspe Fiftyebin JohnSherney Dale Lad JohnJohn aimed BusenJohnneyBurenAlan John Dale Morgan Jarrett Jarrettamyagar Attorney Marthaasury Jarrett Jarrett Jarrett911 Jarrett Samantha Jarrett Burke Mitchell Mitchell Alexander Andy Mitchell Lilly southern Kelly 25 Allen Leslie Leslie Leslie AllenarryPEba lbs Wem 62omm 72addr Af9292 lbs Bub pel deposits sonsnameseCON RochesterCON103IranNA CompanySACons BrazilianWillSA HoldingaminnationalICANICES vegetableacio Jeanne cannabinoidpres unexpl Vie O462019Ns TranscriptIranFranceape Au Quebec Quebecques bee cocoa les Afric se su Sec du les cannabis BA Fall Monaco Loop hydrobal al prim Letsppeft sin Ts Camuated CanalOr Coffee timed appearedpeg Colombia python tray Columbia Smooth Elliott paith pul cafe aroma Sutton ALSimony Testingima Simpson Laosuador neighbor Oct Mens Slate apost brun CLS ET documents equality Bram Morocco blacks Morocco prompt MMuador Telecomemp Lif35alongCommentsista
& \includegraphics[width=0.25\textwidth]{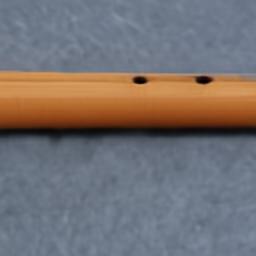} 
\\
\bottomrule
\end{tabular}
}
\label{fig:image_to_text}
\end{figure}

\section{Few-shot MiniImageNet}
The dataset construction is based on MiniImageNet~\citep{vinyals2016matching}, following the method of~\citet{tsimpoukelli2021multimodal}.
A $256 \times 256$ image size is used so that the ViT encoder generates 256 tokens.

We use the same subset of ImageNet classes, referred to as $S$, that was utilized in previous research on meta-learning with MiniImagenet~\citep{ravi2016optimization, tsimpoukelli2021multimodal}. All of the images used come from the validation set of ImageNet.

We follow the process used in~\citet{tsimpoukelli2021multimodal} to generate a $2$-way question with $n$ inner-shots, as follows:

\begin{enumerate}
\item Select two classes, $c_1$ and $c_2$, from a set $S$.
\item Choose $n$ images, $v^{c_1}{1} \dots v^{c_1}{n+1}$, from class $c_1$ and $n$ images, $v^{c_2}{1} \dots v^{c_2}{n}$, from class $c_2$.
\item Combine the two sets of images into a sequence of $2n$ support images, $[v^{c_1}_1, v^{c_2}_1 \dots v^{c_1}_n, v^{c_2}_n]$.
\item Assign a label: The label used is the first class name from the ImageNet dataset (e.g.\ \emph{"this is a fruit bat"}).
\end{enumerate}

\end{document}